\documentclass[runningheads]{llncs}
\usepackage[T1]{fontenc}
\usepackage{graphicx}
\usepackage{ upgreek }
\usepackage{xcolor}
\usepackage{amsmath}
\usepackage{amsfonts}
\usepackage[belowskip=-10pt,aboveskip=0pt]{caption}

\newcommand{\ie}{i.e.,~}
\newcommand{\wrt}{with respect to~}

\newcommand{\x}{\mathbf{x}}
\newcommand{\y}{\mathbf{y}}
\newcommand{\z}{\mathbf{z}}

\newcommand{\X}{\mathbf{x}}
\newcommand{\Y}{\mathbf{y}}
\newcommand{\Z}{\mathbf{z}}

\newcommand{\params}{\Theta}

\newcommand{\mcX}{\mathcal{X}}
\newcommand{\mcY}{\mathcal{Y}}
\newcommand{\mcZ}{\mathcal{Z}}
\newcommand{\mcD}{\mathcal{D}}

\newcommand{\xn}{\x_n}
\newcommand{\yn}{\y_n}
\newcommand{\zn}{\z_n}

\renewcommand{\xi}{\x_i}

\newcommand{\R}[1]{\mathbb{R}^{#1}}

\renewcommand{\L}{\mathcal{L}}

\newcommand{\N}{\mathcal{N}}

\newcommand{\KL}[2]{\operatorname{KL}\left[ #1 \| #2 \right]}
\newcommand{\E}[2]{\mathbb{E}_{#1}\left[#2\right]}

\begin{document}
%
\title{Fully Bayesian VIB-DeepSSM}
\titlerunning{Fully Bayesian VIB-DeepSSM}
\author{Jadie Adams\inst{1,2} \and
Shireen Elhabian\inst{1,2}}
%
\authorrunning{Adams and Elhabian}
%
\institute{Scientific Computing and Imaging Institute, University of Utah, UT, USA \and
Kahlert School of Computing, University of Utah, UT, USA \\
\email{ jadie.adams@utah.edu, shireen@sci.utah.edu }
}
\maketitle              
\begin{abstract}
\setcounter{footnote}{0}
Statistical shape modeling (SSM) enables population-based quantitative analysis of anatomical shapes, informing clinical diagnosis. Deep learning approaches predict correspondence-based SSM directly from unsegmented 3D images but require calibrated uncertainty quantification, motivating Bayesian formulations. Variational information bottleneck DeepSSM (VIB-DeepSSM) is an effective, principled framework for predicting probabilistic shapes of anatomy from images with aleatoric uncertainty quantification. However, VIB is only half-Bayesian and lacks epistemic uncertainty inference. We derive a fully Bayesian VIB formulation and demonstrate the efficacy of two scalable implementation approaches: concrete dropout and batch ensemble. Additionally, we introduce a novel combination of the two that further enhances uncertainty calibration via multimodal marginalization. Experiments on synthetic shapes and left atrium data demonstrate that the fully Bayesian VIB network predicts SSM from images with improved uncertainty reasoning without sacrificing accuracy. \footnote{Source code is publicly available: \url{https://github.com/jadie1/BVIB-DeepSSM}}
\end{abstract}

\section{Introduction}

Statistical Shape Modeling (SSM) is a powerful tool for describing anatomical shapes (\ie bones and organs) in relation to a cohort of interest. 
Correspondence-based shape modeling is popular due to its interpretable shape representation using landmarks or points on anatomical surfaces that are spatially consistent across the population. 
Specifically, each shape is represented by a dense set of correspondences, denoted as a point distribution model (PDM), that is automatically defined on shapes  (e.g., via optimization \cite{cates2017shapeworks} or pairwise parameterization \cite{styner2006spharm}) segmented from 3D medical images.
Conventional SSM pipelines require expert-driven, intensive steps such as segmentation, shape registration, and tuning correspondence optimization parameters or defining an atlas/template for pairwise surface matching. 
Deep learning approaches have mitigated this overhead by providing end-to-end solutions, predicting PDMs from unsegmented 3D images with little preprocessing \cite{bhalodia2018deepssm,uncertaindeepssm,tothova2018uncertainty,tothova2020,adams2022images}.
Such solutions cannot be safely deployed in sensitive, clinical decision-making scenarios without uncertainty reasoning \cite{FDA}, which provides necessary insight into the degree of model confidence and serves as a metric of prediction reliability. 
There are two primary forms of uncertainty aleatoric (or data-dependent) and epistemic (or model-dependent) \cite{kendall2017uncertainties}. 
The overall prediction uncertainty is the sum of the two. 
It is essential to distinguish between these forms, as epistemic is reducible and can be decreased given more training data or by refining the model \cite{der2009aleatory}.
Bayesian deep learning frameworks automatically provide epistemic uncertainty and can be defined to predict distributions, providing aleatoric uncertainty quantification \cite{blundell2015weight,kendall2017uncertainties,maddox2019swag}. 

DeepSSM \cite{bhalodia2018deepssm} is a state-of-the-art framework providing SSM estimates that perform statistically similarly to traditional SSM methods in downstream tasks \cite{deepssm-afib}.  
Uncertain-DeepSSM \cite{uncertaindeepssm} adapted the DeepSSM network to be Bayesian, providing both forms of uncertainties.
DeepSSM, Uncertain-DeepSSM, and other formulations \cite{tothova2018uncertainty} rely on a shape prior in the form of a supervised latent encoding pre-computed using principal component analysis (PCA).
PCA supervision imposes a linear relationship between the latent and the output space,  restricts the learning task, and does not scale in the case of large sets of high-dimensional shape data.
VIB-DeepSSM \cite{adams2022images} relaxes these assumptions to provide improved accuracy and aleatoric uncertainty estimates over the existing state-of-the-art methods \cite{bhalodia2018deepssm,uncertaindeepssm,tothova2018uncertainty}.
This probabilistic formulation utilizes a variational information bottleneck (VIB) \cite{deepvib} architecture to learn the latent encoding in the context of the task, resulting in a more scalable, flexible model
VIB-DeepSSM is self-regularized via a latent prior, increasing generalizability and helping alleviate the need for the computationally expensive DeepSSM data augmentation process.
However, this approach does not quantify epistemic uncertainty because VIB is only half-Bayesian \cite{deepvib}.
In this paper, we propose to significantly extend the VIB-DeepSSM framework to be fully Bayesian, predicting probabilistic anatomy shapes directly from images with both forms of uncertainty quantification. 

The contributions of this work include the following: 
(1) We mathematically derive fully Bayesian VIB from the variational inference perspective.
(2) We demonstrate two scalable approaches for Bayesian VIB-DeepSSM with epistemic uncertainty quantification (concrete dropout and batch ensemble) and compare them to naive ensembling.
(3) We introduce and theoretically justify a novel combination of concrete dropout and ensembling for improved uncertainty calibration.
(4) We illustrate that the fully Bayesian formulations improve uncertainty reasoning (especially the proposed method) on synthetic and real data without sacrificing accuracy.
\section{Background}
We denote a set of paired training data as $\mathcal{D} = \{ \mcX, \mcY \}$. 
$\mcX = \{\xn\}_{n=1}^N$ is a set of $N$ unsegmented images, where $ \xn \in \mathbb{R}^{H \times W \times D}$.
$\mcY = \{\yn\}_{n=1}^N$ is the set of PDMs comprised of $M$ 3D correspondence points, where $\yn \in \mathbb{R}^{3M}$.
VIB utilizes a stochastic latent encoding $\mcZ = \{\zn \}_{n=1}^{N}$, where $\zn \in \R{L}$ and $L \ll 3M$. 

In\textbf{ Bayesian modeling}, model parameters $\params$ are obtained by maximizing the likelihood $p(\y|\x, \params)$.
The predictive distribution is found by marginalizing over $\params$, which requires solving for the posterior $p(\params|\mcD)$. In most cases, $p(\params|\mcD)$ is not analytically tractable; thus, an approximate posterior $q(\params)$ is found via variational inference (VI).
Bayesian networks maximize the VI evidence lower bound (ELBO) by minimizing:
\begin{equation}
    \L_\text{VI}= \E{\tilde{\params} \sim q(\params)}{ - \log p(\Y | \X, \tilde{\params})} + \beta \KL{q(\params)}{ p(\params) }
    \label{eq:BNN}
\end{equation}
where $p(\params)$ is the prior on network weights, and $\beta$ is a weighting parameter.

The deep \textbf{Variational Information Bottleneck} (VIB) \cite{deepvib} model learns to predict $\y$ from $\x$ using a low dimensional stochastic encoding $\z$.
The VIB architecture comprises of a stochastic encoder parameterized by $\phi$, $q(\z|\x, \phi)$, and a decoder parameterized by $\theta$, $p(\y|\z, \theta)$ (Figure \ref{fig:arch}).
VIB utilizes VI to derive a theoretical lower bound on the information bottleneck objective:
\begin{equation}
    \L_{VIB} = \E{\hat{\Z} \sim q(\Z|\X, \phi)}{- \log p(\Y|\hat{\Z}, \theta)} + \beta\KL{q(\Z|\X, \phi)}{p(\Z)}
    \label{eq:VIB_elbo}
\end{equation}
The entropy of the $p(\y|\z)$ distribution (computed via sampling) captures aleatoric uncertainty.
The VIB objective has also been derived using an alternative motivation: Bayesian inference via optimizing a PAC style upper bound on the true negative log-likelihood risk \cite{alemi2020vib}. 
Through this PAC-Bayes lens, it has been proven that VIB is \textit{half Bayesian}, as the Bayesian strategy is applied to minimize an upper bound with respect to the conditional expectation of $\Y$, but Maximum Likelihood Estimation (MLE) is used to approximate the expectation over inputs.
The VIB objective can be made a fully valid bound on the true risk by applying an additional PAC-Bound with respect to the parameters, resulting in a fully Bayesian VIB that captures epistemic uncertainty in addition to aleatoric.

\section{Methods}

\subsection{Bayesian Variational Information Bottleneck}

In fully-Bayesian VIB (BVIB), rather than fitting the model parameters $\params = \{\phi, \theta\}$ via MLE, we use VI to approximate the posterior $p(\params|\mcD)$. There are now two intractable posteriors $p(\Z|\X,\phi)$ and $p(\params|\X, \Y)$. The first is approximated via $q(\Z|\X, \phi)$ as in Eq. \ref{eq:VIB_elbo} and the second is approximated by $q(\params)$ as in Eq. \ref{eq:BNN}. Minimizing these two KL divergences via a joint ELBO gives the objective (see Appendix A) for derivation details):

\begin{equation}
    \L_\text{BVIB} = \E{\tilde{\params}}{\E{\hat{\Z}}{- \log p(\Y|\hat{\Z}, \tilde{\theta})} + \KL{q(\Z|\X, \tilde{\phi})}{p(\Z)}} + \KL{q(\params)}{p(\params)}
    \label{eq:BVIB_loss}
\end{equation}
where $ \tilde{\params} \sim q(\params)$ and $ \hat{\Z} \sim q(\Z|\X, \tilde{\phi})$.
This objective is equivalent to the BVIB objective acquired via applying a PAC-Bound \wrt the conditional expectation of targets and then another \wrt parameters \cite{alemi2020vib}. This is expected, as it has been proven that the VI formulation using ELBO and the PAC-Bayes formulation with negative log-likelihood as the risk metric are algorithmically identical \cite{thakur2019unifying}. Additionally, this matches the objective derived for the Bayesian VAE when $\Y=\X$ \cite{BVAE}.
Implementing BVIB requires defining a prior distribution for the latent representation $p(\Z)$ and the network weights $p(\params)$. Following VIB, we define, $p(\z) = \N(\z|\mathbf{0},\mathbb{I})$. Different methods exist for defining $p(\params)$, and multiple approaches are explored in the following section.

\subsection{Proposed BVIB-DeepSSM Model Variants}

In adapting VIB-DeepSSM to be fully Bayesian, we propose utilizing two approaches that have demonstratively captured epistemic uncertainty without significantly increasing computational and memory costs: concrete dropout \cite{gal2017concrete} and batch ensemble \cite{wen2020batchensemble}. Additionally, we propose a novel integration for a more flexible, multimodal posterior approximation.

\textbf{Concrete Dropout} (CD) utilizes Monte Carlo dropout sampling as a scalable solution for approximate VI \cite{gal2017concrete}. Epistemic uncertainty is captured by the spread of predictions with sampled dropout masks in inference. CD automatically optimizes layer-wise dropout probabilities along with the network weights. 

\textbf{Naive Ensemble} (NE) models combine outputs from several networks for improved performance. Networks trained with different initialization converge to different local minima, resulting in test prediction disagreement\cite{fort2019deep}. The spread in predictions effectively captures epistemic uncertainty \cite{rahaman2021uncertainty}.
NE models are computationally expensive, as cost increases linearly with number of members.

\textbf{Batch Ensemble} (BE) \cite{wen2020batchensemble} compromises between a single network and NE, balancing the trade-off between accuracy and running time and memory. 
In BE, each weight matrix is defined to be the Hadamard product of a shared weight among all ensemble members and a rank-one matrix per member. 
BE provides an ensemble from one network, where the only extra computation cost is the Hadamard product, and the only added memory overhead is sets of 1D vectors.

\textbf{Novel Integration of Dropout and Ensembling:}
Deep ensembles have historically been considered a non-Bayesian competitor for uncertainty estimation.
However, recent work argues that ensembles approximate the predictive distribution more closely than canonical approximate inference procedures (\ie VI) and are an effective mechanism for approximate Bayesian marginalization \cite{wilson2020case}.
Combining traditional Bayesian methods with ensembling improves the fidelity of approximate inference via multimodal marginalization, resulting in a more robust, accurate model \cite{wilson2020bayesian}.
In concrete dropout, the approximate variational distribution is parameterized via a concrete distribution. While this parameterization enables efficient Bayesian inference, it greatly limits the expressivity of the approximate posterior. To help remedy this, we propose integrating concrete dropout and ensembling (BE-CD and NE-CD) to acquire a multimodal approximate posterior on weights for increased flexibility and expressiveness.
While ensembling has previously been combined with MC dropout for regularization\cite{wen2020batchensemble}, this combination has not been proposed with the motivation of multimodal marginalization for improved uncertainty calibration. 

\begin{figure}[!ht]
    \begin{center}
    \includegraphics[width=\textwidth]{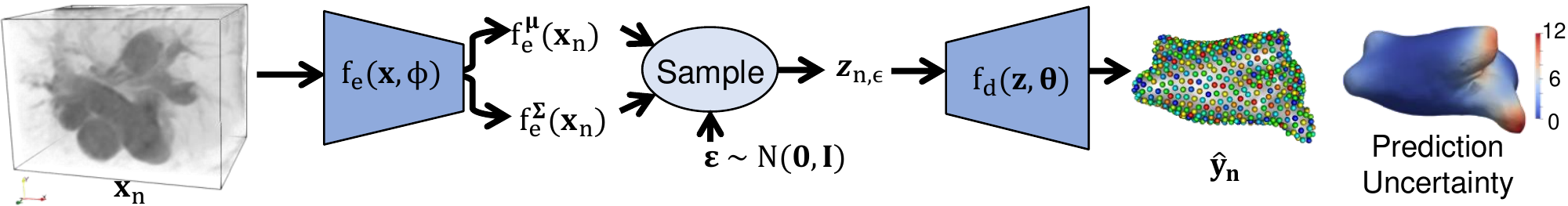}
    \end{center}
    \caption{\textbf{Common VIB-DeepSSM Architecture} for all proposed variants.}
    \label{fig:arch}
\end{figure} 

\subsection{BVIB-DeepSSM Implementation}
We compare the proposed BVIB approaches with the original VIB-DeepSSM formulation \cite{adams2022images}.
All models have the overall structure shown in Figure \ref{fig:arch}, comprised of a 3D convolutional encoder($f_e$) and fully connected decoder ($f_d$).
CD models have concrete dropout following every layer, BE weights have four members (the maximum GPU memory would allow), and four models were used to create NE models for a fair comparison.
Following \cite{adams2022images}, burn-in is used to convert the loss from deterministic (L2) to probabilistic (Eqs 10, \ref{eq:BVIB_loss}, 13) \cite{adams2022images}. 
This counteracts the typical reduction in accuracy that occurs when a negative log-likelihood based loss is used with a gradient-based optimizer \cite{seitzer2021pitfalls}. An additional dropout burn-in phase is used for CD models to increase the speed of convergence\cite{adams2022images}.
All models were trained until the validation accuracy had not decreased in 50 epochs. A table of model hyperparameters and tested ranges is proved in Appendix C.
The training was done on Tesla V100 GPU with Xavier initialization \cite{xavier2010initialization}, Adam optimization \cite{adam}.
The prediction uncertainty is a sum of the epistemic (variance resulting from marginalizing over $\Theta$) and aleatoric (variance resulting from marginalizing over $\Z$) uncertainty (see Appendix B for calculation details).

\section{Results}
We expect well-calibrated prediction uncertainty to correlate with the error, aleatoric uncertainty to correlate with the input image outlier degree (given that it is data-dependent), and epistemic uncertainty to correlate with the shape outlier degree (i.e., to detect out-of-distribution data). 
The outlier degree value for each mesh and image is quantified by running PCA (preserving $95\%$ of variability) and then considering the Mahalanobis distance of the PCA scores to the mean (within-subspace distance) and the reconstruction error (off-subspace distance). The sum of these values provides a measure of similarity to the whole set in standard deviation units \cite{moghaddam1997probabilistic}.
Experiments are designed to evaluate this expected correlation as well as accuracy, which is calculated as the root mean square error (RMSE) between the true and predicted points. Additionally, we quantify the \textit{surface-to-surface distance} between a mesh reconstructed from the predicted PDM (predicted mesh) and the ground truth segmented mesh. The reported results are an average of four runs for each model, excluding the NE models, which ensemble the four runs.

\subsection{Supershapes Experiments}

Supershapes (SS) are synthetic 3D shapes parameterized by variables that determine the curvature and number of lobes \cite{gielis2003generic}. We generated 1200 supershapes with lobes randomly selected between 3 and 7 and curvature parameters randomly sampled from a $\chi^2$ distribution with 4 degrees of freedom. 
Corresponding 3D images were generated with foreground and background intensity values modeled as Gaussian distributions with different means and equal variance. Images were blurred with a Gaussian filter (size randomly selected between 1 and 8) to mimic diffuse shape boundaries. 

\begin{figure}[!h]
    \begin{center}
    \includegraphics[width=.9\textwidth]{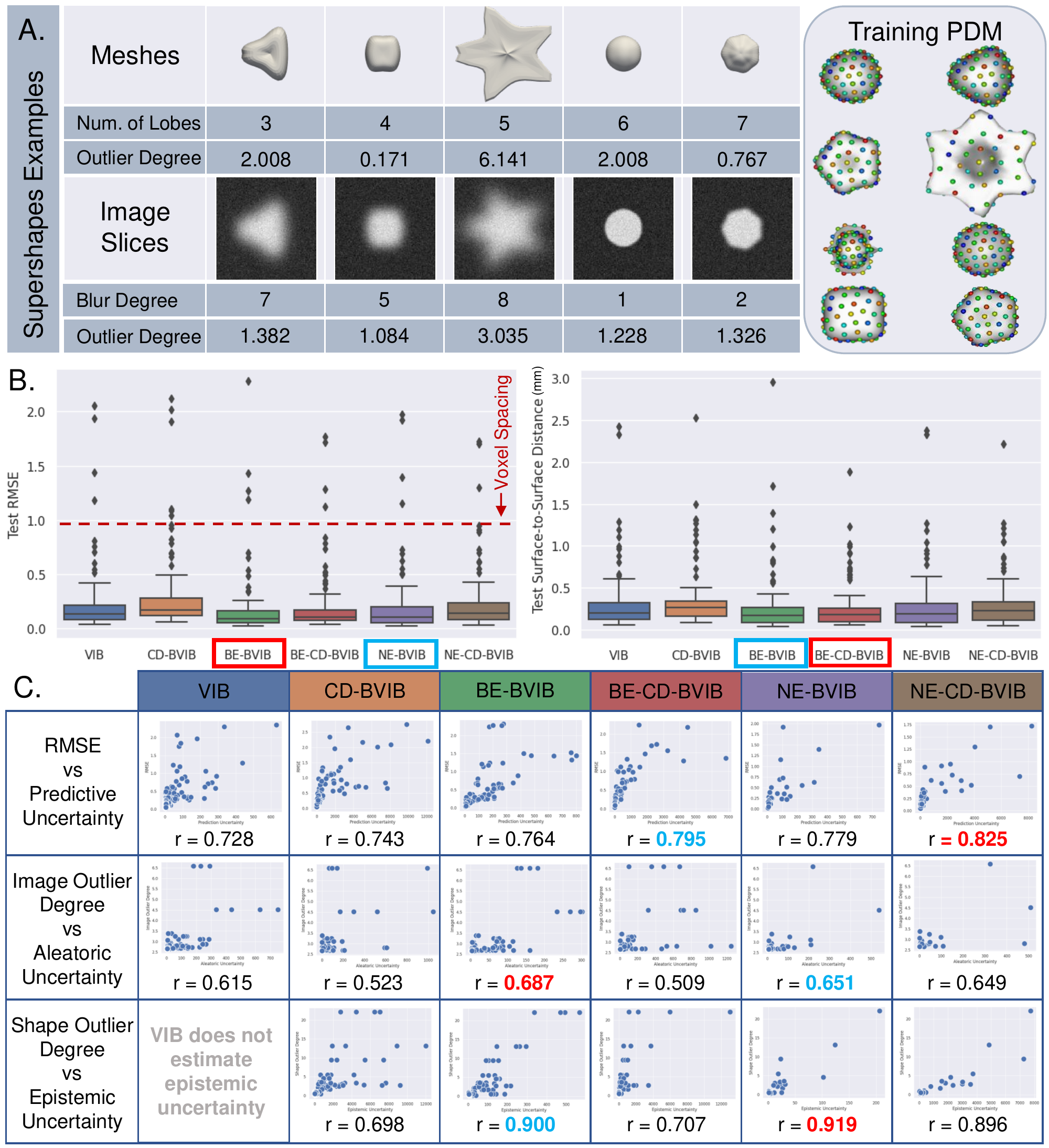}
    \end{center}
    \caption{\textbf{Supershapes} (A) Left: Five examples of SS mesh and image pairs with corresponding outlier degrees. Right: Examples of training points, where color denotes point correspondence. (B) Distribution of errors over the test set, lower is better. (C) Uncertainty correlation, where a higher Pearson r coefficient suggests better calibration. Best values are marked in red, and the second best in blue.}
    \label{fig:SS_combo}
\end{figure}  

Figure \ref{fig:SS_combo}A displays example shape meshes and images with corresponding outlier degrees, demonstrating the wide variation. 
We randomly split the mesh-image pairs to create a training set of size 1000, a validation set of size 100, and a testing set of size 100. 
ShapeWorks \cite{cates2017shapeworks} was used to optimize PDMs of 128 points on the training set. Target PDMs were then optimized for validation and test sets, keeping the training PDMs fixed so that the test set statistics were not captured by the training PDMs. 

Figure \ref{fig:SS_combo}B demonstrates that all BVIB models performed similarly or better than the baseline VIB in terms of RMSE and surface-to-surface distance, with the BE models performing best.
Interestingly, the BE models were more accurate than the NE. This effect could result from the random sign initialization of BE fast weights, which increases members diversity.
Adding CD hurt the accuracy slightly, likely because the learning task is made more difficult when layer-wise dropout probabilities are added as variational parameters. However, CD is the cheapest way to add epistemic uncertainty and improve prediction uncertainty calibration. 
Figure \ref{fig:SS_combo}C demonstrates prediction uncertainty is well-calibrated for all models (with an error correlation greater than 0.7) and NE-CD-BVIB achieves the best correlation. The aleatoric and epistemic uncertainty correlation was similar across models, with the ensemble-based models performing best.

\subsection{Left Atrium Experiments}

The left atrium (LA) dataset comprises 1041 anonymized LGE MRIs from unique patients. The images were manually segmented  at the University of Utah Division of Cardiovascular Medicine with spatial resolution $0.65 \times 0.65 \times 2.5$ mm$^3$, and the endocardium wall was used to cut off pulmonary veins. 
The images were cropped around the region of interest, then downsampled by a factor of 0.8 for memory purposes. 
This dataset contains significant shape variations, including overall size, LA appendage size, and pulmonary veins' number and length. The input images vary widely in intensity and quality, and LA boundaries are blurred and have low contrast with the surrounding structures. 
Shapes and image pairs with the largest outlier degrees were held out as outlier test sets, resulting in a \textit{shape outlier} test set of 40 and \textit{image outlier} test set of 78. 
We randomly split the remaining samples (90\%, 10\%, 10\%) to get a training set of 739, a validation set of 92, and an \textit{inlier} test set of 92. The target PDMs were optimized with ShapeWorks \cite{cates2017shapeworks} to have 1024 particles.

\begin{figure}[ht!]
    \begin{center}
    \includegraphics[width=.9\textwidth]{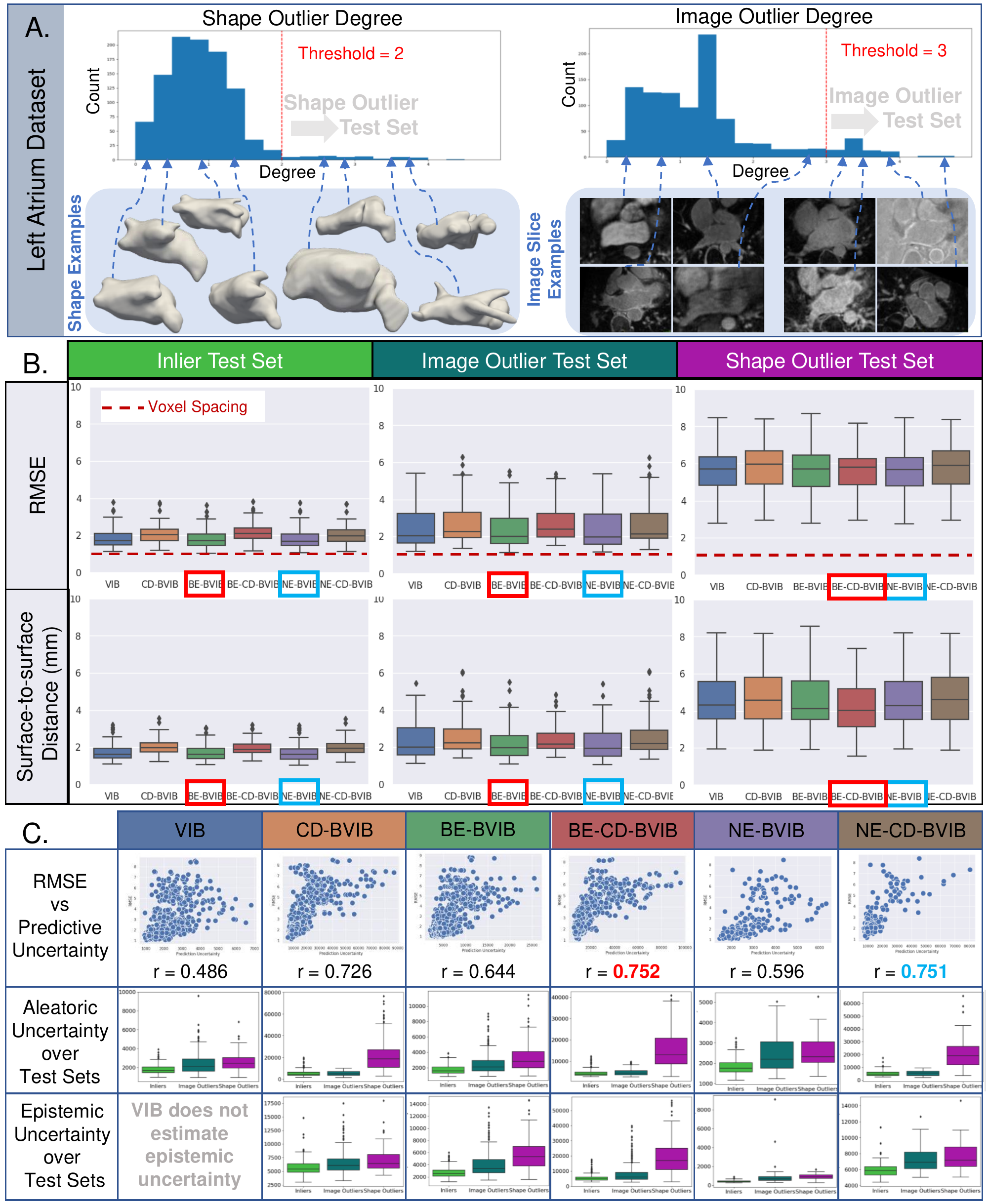}
    \end{center}
    \caption{\textbf{Left Atrium} (A) The distribution of shape and image outlier degrees with thresholds is displayed with examples. (B) Box plots show the distribution of errors over the test sets. (C) Scatterplots show uncertainty correlation with error across test sets and box plots show the distribution of uncertainty for each test set. The best values are marked in red, and the second best in blue.}
    \label{fig:LA_combo}
\end{figure}

The accuracy and uncertainty calibration analysis in Figure \ref{fig:LA_combo}B and \ref{fig:LA_combo}C show similar results to the supershapes experiment. 
In both experiments, the proposed combination of dropout and ensembling provided the best-calibrated prediction uncertainty, highlighting the benefit of multimodal Bayesian marginalization. Additionally, the proposed combination gave more accurate predictions on the LA outlier test sets, suggesting improved robustness. 
BE-CD-BVIB provided the best prediction uncertainty for the LA and the second best (just behind NE-CD-BVIB) for the SS. 
BE-CD-BVIB is a favorable approach as it does not require training multiple models as NE does and requires relatively low memory addition to the base VIB model. 
Further qualitative LA results are provided in Appendix F in the form of heat maps of the error and uncertainty on test meshes. Here we can see how the uncertainty correlates locally with the error. As expected, both are highest in the LA appendage and pulmonary veins region, where LA's and the segmentation process vary the most. 
It is worth noting a standard normal prior was used for $p(\Z)$ in all models. Defining a more flexible prior, or potentially learning the prior, could provide better results and will be considered in future work.

\section{Conclusion} 
The traditional computational pipeline for generating Statistical Shape Models (SSM) is expensive and labor-intensive, which limits its widespread use in clinical research. Deep learning approaches have the potential to overcome these barriers by predicting SSM from unsegmented 3D images in seconds, but such a solution cannot be deployed in a clinical setting without calibrated estimates of epistemic and aleatoric uncertainty. The VIB-DeepSSM model provided a principled approach to quantify aleatoric uncertainty but lacked epistemic uncertainty. To address this limitation, we proposed a fully Bayesian VIB model that can predict anatomical SSM with both forms of uncertainty. We demonstrated the efficacy of two practical and scalable approaches, concrete dropout and batch ensemble, and compared them to the baseline VIB and naive ensembling. Finally, we proposed a novel combination of dropout and ensembling and showed that the proposed approach provides improved uncertainty calibration and model robustness on synthetic supershape and real left atrium datasets. While combining Bayesian methods with ensembling increases memory costs, it enables multimodal marginalization improving accuracy. These contributions are an important step towards replacing the traditional SSM pipeline with a deep network and increasing the feasibility of fast, accessible SSM in clinical research and practice.

\subsubsection{Acknowledgements}
This work was supported by the National Institutes of Health under grant numbers NIBIB-U24EB029011, NIAMS-R01AR076120, \\ NHLBI-R01HL135568, and  NIBIB-R01EB016701.
The content is solely the responsibility of the authors and does not necessarily represent the official views of the National Institutes of Health.
The authors would like to thank the University of Utah Division of Cardiovascular Medicine for providing left atrium MRI scans and segmentations from the Atrial Fibrillation projects.

%
%
%
\bibliographystyle{splncs04}
\bibliography{refs}
\appendix

\section{Bayesian VIB Derivation}\label{app:BVIB}
\vspace{-0.1in}

In Bayesian inference, for a given test sample, $\x^*$, the predictive distribution is found by marginalizing over $\params$:
$
    p(\y^*|\x^*, \mcX, \mcY) = \int  p(\y^*|\x^*, \params)p(\params|\mcX, \mcY)d\params
    \label{eq:BNN_pred}
$.
In VIB, for a given test sample $\x^*$, the predictive distribution is defined as:
$
    p(\y^*|\x^*, \params) = \int p(\y^*|\z^*, \theta)p(\z^*|\x^*,\phi)d\Z
    \label{eq:VIB_pred}
$.
To extend VIB to be fully Bayesian, we place a prior ($p(\params)$) over  model parameters ($\params = \{\phi, \theta\}$) and approximate the posterior ($p(\params|\mcD)$).
Under this framework, we combine the predictive distributions as follows:
\begin{equation}
        p(\y^*|\x^*, \mcX, \mcY) = \int \left( \int p(\y^*|\z^*, \theta)p(\z^*|\x^*,\phi)d\Z \right) p(\params|\mcX, \mcY)d\z d\params 
\end{equation}
Here we have two intractable posteriors $p(\Z|\X,\phi)$ and $p(\params|\X, \Y)$, thus we apply VI twice. The first is approximated via $q(\Z|\X, \phi)$ as is done in Equation \ref{eq:VIB_elbo} and the second is approximated by $q(\params)$ as is done in Equation \ref{eq:BNN}. Directly combining Equations \ref{eq:BNN} and \ref{eq:VIB_elbo} results in the BVIB objective shown in Equation \ref{eq:BVIB_loss}.

\section{Uncertainty Calculation}
\label{app:UQ}
\vspace{-0.1in}

To estimate epistemic uncertainty in BVIB-DeepSSM, we generate multiple predictions $(\hat{\y}_n, \mathbf{\hat{\sigma}}^2_n)$ with  $T$ weights sampled from the approximate posterior. In CD, $T$ is the number of different dropout masks used in generating predictions (30). In BE and NE, $T$ is the number of ensemble members (4). In the proposed CD ensemble integration, $T$ is the number of dropout masks times the number of ensemble members (120).
The prediction uncertainty is expressed as the sum of the aleatoric and epistemic uncertainty \cite{kendall2017uncertainties}: 
\begin{equation}
    \mathbf{\Sigma}^2_n = \frac{1}{T}\sum_{t=1}^T  \hat{\y}_{n,t}^2 -  \left(\frac{1}{T}\sum_{t=1}^T \hat{\y}_{n,t}\right)^2 + \frac{1}{T}\sum_{t=1}^T \hat{\mathbf{\sigma}}_{n,t}^2
\end{equation}

\section{Model Parameters}
Architecture: The encoder, $f_e$ is comprised of 3D convolution, 3D batch normalization, and fully connected (FC) layers as in \cite{bhalodia2018deepssm}. The decoder is comprised of three FC layers and Parametric ReLU activation \cite{he2015rectifiers}.
\vspace{-0.3in}
\label{app:params}
\begin{table}[h]
\caption{Parameter values used in all experiments with the tested range.}
\centering
\resizebox{0.75\textwidth}{!}{%
\begin{tabular}{|l|c|c|}
\hline
\textbf{Parameter} & \textbf{Selected Value} & \textbf{Tested Range/Explanation} \\ \hline
$\beta$ (Equations \ref{eq:VIB_elbo} and \ref{eq:BVIB_loss})& 0.01 & $[1e^{-8}, 0.9]$  \\ \hline
 Fixed Learning Rate & $5e^{-5}$  &  $[1e^{-6}, 1e^{-3}]$ \\ \hline
CD Length Scale &  $1e^{-3}$  &  $[1e^{-8}, 1e^{2}]$ \\ \hline
Batch Size &  6  &  Maximum memory allowed. \\ \hline
Number of Ensemble Members &  4  &  Maximum memory allowed. \\ \hline
Adam Optimization Betas &  (0.9, 0.999)  &  Default used. \\ \hline
Adam Optimization Eps &  $1e^{-8}$  &  Default used. \\ \hline
Adam Optimization Weight Decay &  0  &  Default used. \\ \hline
\end{tabular}%
}
\end{table}

\section{Example Left Atrium Results}
\label{app:LA}

\begin{figure}[!ht]
    \begin{center}
    \includegraphics[width=.8\textwidth]{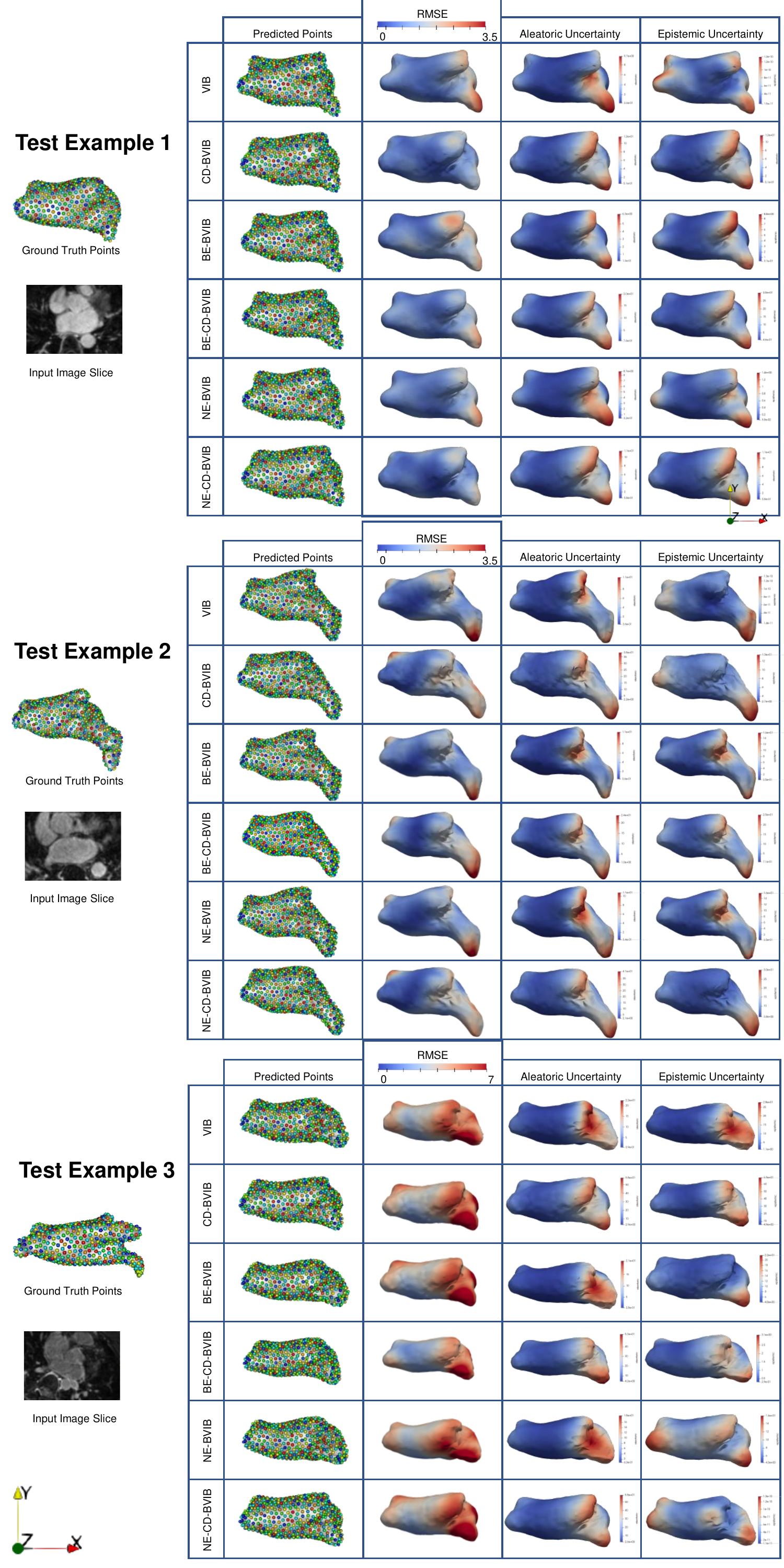}
    \end{center}
    \caption{Three examples from the LA test set are provided, including input image slices, ground truth points, and the model output point and uncertainty predictions. Example 1 demonstrates a typical case, example 2 contains an enlarged LA appendage, and example 3 has multiple long pulmonary veins. These examples show the local calibration of the uncertainty estimation, demonstrating the utility of uncertainty quantification in clinical analysis. }
    \label{fig:ex1}
\end{figure}

\end{document}